\documentclass{article}

 \PassOptionsToPackage{numbers, compress}{natbib}

\usepackage[final]{tackling_climate_workshop_style}

\usepackage[utf8]{inputenc} 
\usepackage[T1]{fontenc}    
\usepackage{hyperref}       
\usepackage{url}            
\usepackage{booktabs}       
\usepackage{amsfonts}       
\usepackage{nicefrac}       
\usepackage{microtype}      
\usepackage{graphicx}
\usepackage{multirow}
\usepackage[table,xcdraw]{xcolor}
\usepackage{todonotes}
\usepackage{tabularx}
\usepackage{enumitem}
\usepackage{wrapfig}
\usepackage{subcaption}

\title{\textit{Climate Policy Tracker}: Pipeline for automated analysis of public climate policies}

\author{%
  Artur Żółkowski\\
  Warsaw University of Technology\\
  \texttt{artur.zolkowski.stud@pw.edu.pl} \\
   \And
   Mateusz Krzyziński \\
  Warsaw University of Technology \\
   \texttt{mateusz.krzyzinski.stud@pw.edu.pl} \\
  \And
   Piotr Wilczyński \\
  Warsaw University of Technology \\
   \texttt{piotr.wilczynski.stud@pw.edu.pl} \\
   \And
  Stanisław Giziński \\
  University of Warsaw \\
   \texttt{s.gizinski@student.uw.edu.pl} \\
  \And
   Emilia Wiśnios\thanks{Corresponding author} \\
    University of Warsaw \\
   \texttt{e.wisnios@student.uw.edu.pl} \\
   \And
   Bartosz Pieliński \\
   University of Warsaw \\
   \texttt{b.pielinski@uw.edu.pl}
   \And 
   Julian Sienkiewicz \\
   Warsaw University of Technology \\
   \texttt{julian.sienkiewicz@pw.edu.pl} 
   \And 
   Przemysław Biecek \\
   Warsaw University of Technology \\
   \texttt{przemyslaw.biecek@pw.edu.pl}
}

\begin{document}

\maketitle

\begin{abstract}

The number of standardized policy documents regarding climate policy and their publication frequency is significantly increasing. The documents are long and tedious for manual analysis, especially for policy experts, lawmakers, and citizens who lack access or domain expertise to utilize data analytics tools. Potential consequences of such a situation include reduced citizen governance and involvement in climate policies and an overall surge in analytics costs, rendering less accessibility for the public. In this work, we use a Latent Dirichlet Allocation-based pipeline for the automatic summarization and analysis of 10-years of national energy and climate plans (NECPs) for the period from 2021 to 2030, established by 27 Member States of the European Union. We focus on analyzing policy framing, the language used to describe specific issues, to detect essential nuances in the way governments frame their climate policies and achieve climate goals. The methods leverage topic modeling and clustering for the comparative analysis of policy documents across different countries. It allows for easier integration in potential user-friendly applications for the development of theories and processes of climate policy. This would further lead to better citizen governance and engagement over climate policies and public policy research.

\end{abstract}


\section{Introduction}

Machine learning (ML) systems have already been used to analyze climate change~\cite{Rolnick2022}. They are able to follow climate change itself as well as develop and evaluate solutions preventing or even reversing the process. ML in the form of Natural Language Processing (NLP) has also started to be implemented to study discussions around climate change and climate policies~\cite{stede-patz-2021-climate}. It has also been used to research how governments around the world reacted to climate change~\cite{beyond_modeling, nlp_climate_policy}. The system presented in this paper is a part of the latter strand of research. It aims to develop NLP methods dedicated to analyzing specific types of policy documents -- highly standardized national climate strategies and reports.    

In recent years, as the danger of climate change has become acknowledged by public opinion, policymakers, and politicians, the number of policy documents dedicated to climate policy has been steadily increasing (Appendix \ref{appendix:policies}). However, after the introduction of the Paris Agreement, a specific group of documents has started to be produced by national governments of the Member States of the European Union (EU) -- climate strategies and reports on implementing these strategies. The Agreement (Article 13) established an Enhanced Transparency Framework design to provide guidance for regular reporting on national climate policies. As a consequence, the Agreement has stimulated the development of policy documents having similar structures as they address previously agreed issues. Those documents have been analyzed so far mainly through the perspective of climate policy indicators showing commitments of national governments to limit climate change \cite{opportunities_and_gaps, planning_for_net_zero}. 

In this study, we propose the pipeline (Figure \ref{fig:pipeline}) that focuses on different aspects of these documents -- the way they frame climate policy. The issue framing has been found
to influence policy dynamics over the long run \cite{daviter_introduction_2011}. The similar structure of the documents allows us to compare the policy framing in different dimensions separately, which allows for a more fine-grained analysis of policy frames.

The pipeline was created based on two assumptions: (1) the number and publishing frequency of standardized documents regarding climate policy will soon increase significantly; (2) NLP methods allow a relatively quick and comprehensive summary of these documents and their comparison. They can also reveal important nuances in how governments frame their climate policies. 

The main contributions of our work are as follows:
\begin{enumerate}
\itemsep0em
    \item We propose a pipeline for analyzing similarly structured documents regarding climate public policy, leveraging Natural Language Processing.
    \item We provide an easy-to-use \href{https://mi2.ai/climate}{web application} for exploring structured results of our pipeline.
\end{enumerate}

\section{Data}
We analyze the 10-year national energy and~climate plans (NECPs) for~the~period from 2021 to~2030 established by 27 Member States of the EU. These documents contain~information on how individual countries want to~relate to~the~EU's critical energy and~climate policy dimensions. Each plan was published in~English and~the~country's national language on the EU's official website. Only documents written in English were considered in this study. All of the data and information used in the paper, either as a subject of the analysis or to support its claims, are publicly available  \href{https://energy.ec.europa.eu/topics/energy-strategy/national-energy-and-climate-plans-necps_en}{on this website}.

To ensure the parallelism of the plans between the Member States, NECPs had to follow a unified structure facilitating comparison and aggregation. According to the template, five Energy Union dimensions and five sections should be distinguished. See Appendix \ref{appendix:necp_structure} for more details.

We performed preprocessing of all documents; a detailed description of this process is available in Appendix \ref{appendix:dataprep}.

\section{Methodology}

\paragraph{Topic modeling}
For modeling topics, we use the Latent Dirichlet Allocation (LDA) model \cite{blei2003latent} as it assumes each document is a finite mixture over an underlying set of topics, which is in line with the assumptions of our study.

Modeling was carried out using the \texttt{Gensim} package \cite{rehurek2011gensim} separately on seven sub-datasets -- five defined by Energy Union dimensions and~two related to~subsections where these dimensions were not distinguished. In~addition, words were filtered to~include only those appearing in~at least four analyzed texts, which translates into the~presence of~words in~documents submitted by at~least two different countries. Such filtering ensures that texts are not grouped based upon expressions unique to~each country yet unrelated to~the~document's topic.

We assumed that prior over topic-word distribution may be asymmetric and~it was matched automatically based on the~subcorpora. For selecting the~parameters describing the~symmetric prior over document-topic distribution ($\alpha$ parameter in \texttt{Gensim} LDA), we performed two grid searches. We observed that choosing the correct number of topics is more important than choosing the optimal $\alpha$. However, models gave more intuitive and interpretable results for $\alpha \in \{50, 100, 150, 200, 250\}$.  The performance of such parameter values is probably explained by the fact that considered documents are similar in terms of topics and differ only in nuances. We assessed the quality of topic modeling using two coherence measures, CV \cite{cv} and UMass \cite{umass}. Moreover, we manually analyzed and~validated the~results, which is~necessary to~obtain~semantically meaningful topics \cite{topicvalidation}. 

\paragraph{Topic interpretation}
One of~the~most challenging aspects of~topic modeling is~the~characterization of~the~topics learned by the~used model \cite{describingtopics}. Topics obtained using LDA do not have an interpretable description, but usually, the~most common words are used to~describe topics. We used the 25 most popular words for automatic topic naming using GPT-3 \cite{gpt3}. 

However, such a~representation does not give the~complete picture and~can even be misleading, as each topic is~a~distribution over \emph{the~entire} vocabulary. Therefore, we used the~\texttt{LDAvis} tool \cite{sievert-shirley-2014-ldavis} to~characterize and~interpret the~topics. This interactive method provides a~global overview of~the~topics and~illustrates how they differ. At the~same time, it allows for a~scrutinous examination of~the~terms most associated with each topic. Precisely, two metrics were used in~the~topic analysis and~description process: the~saliency metric \cite{saliencymetric} used to~identify the~most informative words, and the~relevance metric \cite{sievert-shirley-2014-ldavis} used to~rank words within~topics and~identify the~most relevant terms. Based on this exploration and automatic suggestions, we made the final naming of topics. 

\paragraph{Comparison of policy frames mixture among countries}
In~the~analysis, we used similarity measures computed in~terms of~similarity of~distributions described directly for the~LDA model \cite{similarities}. We select an information radius measure (also known as Jensen-Shannon divergence).

We examined the~dependencies between the~topic mixtures of~individual countries by grouping them with a~hierarchical clustering algorithm. An agglomerative approach with an~average linkage criterion was used. The~choice of~this clustering procedure was due to~the~fact that it produces a~complete dendrogram showing the~range of~levels at~which countries are related. Thus, it presents a~more holistic picture of~the~agendas' underlying structure than other methods that return a single version of clustering. In~addition, we used the t-distributed stochastic neighbor embedding (t-SNE) \cite{tsne} to~visualize the results of clustering.

We examined how the clusters of countries differ across dimensions. This has a twofold purpose. First, it allows us to validate the clustering and topics against the expert knowledge of well-established countries' alliances, such as the Visegrad Group. Secondly, it allows us to pinpoint other, non-obvious groups of countries with matching or similar policy framings. This creates an opportunity to provide new hypotheses for further geopolitical and climate studies.

\begin{wrapfigure}[16]{r}{0.6\textwidth}
\centering
        \raisebox{0pt}[\dimexpr\height-1.35\baselineskip\relax]{\includegraphics[width=0.59\textwidth]{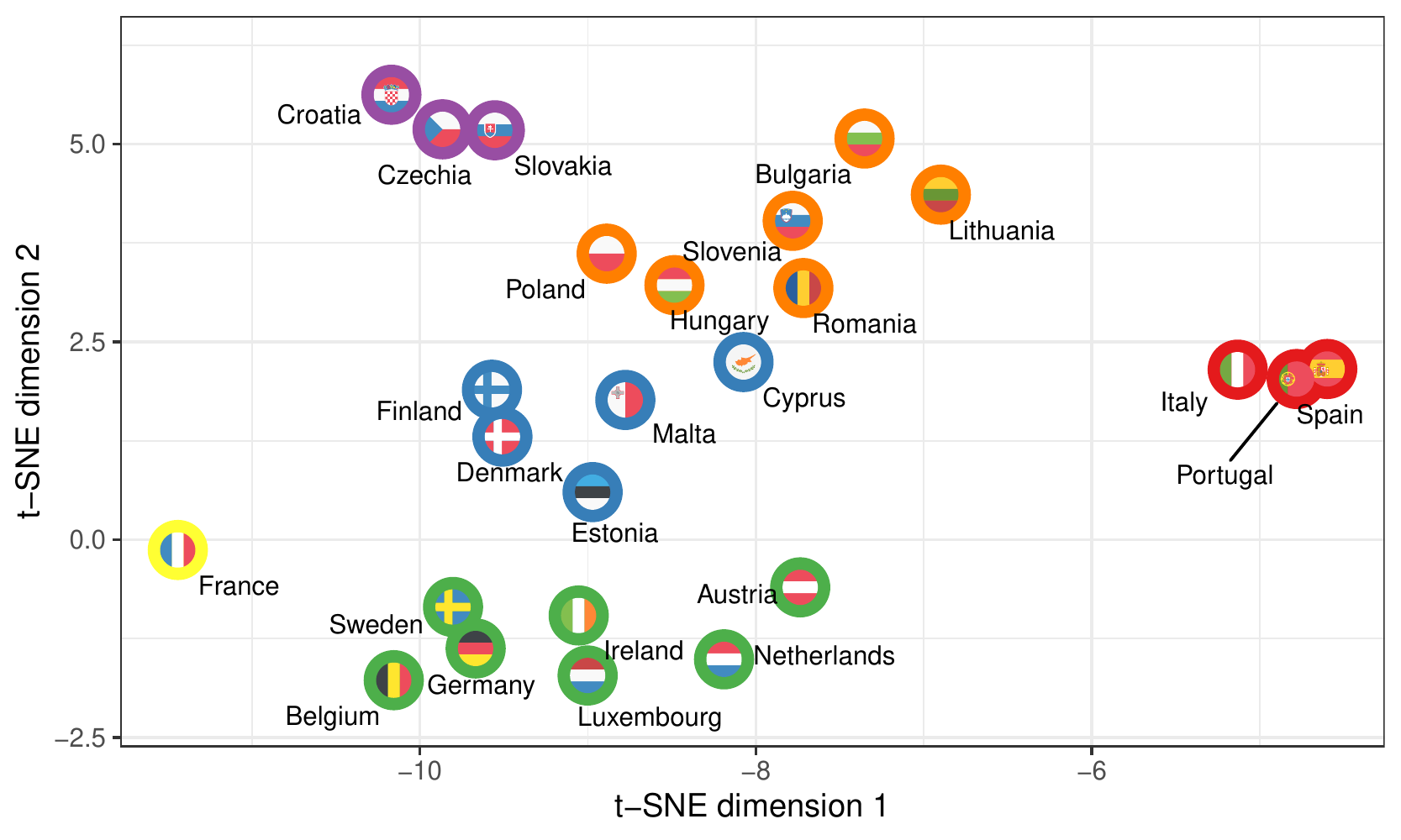}}%
\caption{Visualization of selected country clustering for the \textit{Decarbonisation} dimension. Positions in two-dimensional space are based on t-SNE embedding.}
\label{fig:clustering}
\end{wrapfigure}

\section{Results}

\paragraph{Web application for exploration of the results}
The~result of~topic modeling are~the~distributions of~topics in~each analyzed text. Matrices composed of~these vectors for each of~the~seven models are the~direct result of~the~modeling process, and they can be analyzed in a variety of ways. Thus, we have prepared a demo version of the web application that enables in-depth, interactive exploration on many levels, see Appendix \ref{appendix:app}. First, it is possible to examine the distribution of topics for the selected country and compare them between sections (Figure \ref{fig:radar}) or juxtapose them with the overall results for other countries (Figure \ref{fig:violin}).
Further, the application allows for clustering based on the found representation (Figure \ref{fig:clustering}). The interpretability of the results is ensured by analysis that allows the exploration of particular topics through their keywords (Table \ref{tab:topics}). Another feature is the ability to compare the values for topics with external variables (Figure \ref{fig:compare}).

\paragraph{Differences between sections across dimensions}
The modeling results can also be used for other, more sophisticated analyses, such as comparisons between \textit{National objective and targets} and \textit{Policies and measures} sections to detect inconsistencies in agendas. It turns out that some states have more discrepancies and that this phenomenon is more typical for some dimensions, see Appendix \ref{appendix:inconsistencies}.

\paragraph{Policy frames focus for countries} 
We find there are 5 policy frames for \textit{Decarbonization} and \textit{Energy efficiency}, 3 policy frames for \textit{Energy security}, \textit{Internal market}, and \textit{R\&I and Competitiveness}. Overall, our system detects 19 policy frames across all dimensions. A full list of policy frames is available in Appendix \ref{appendix:policyframes}.
By analyzing the topics' distribution, we conclude which country focuses on which policy frame in each of the dimensions. As an example, we can see that Bulgaria is the country that devotes the most parts of the \textit{Decarbonization} section to the \textit{Greenhouse gas emission}, (.65 topic probability). Netherlands is an outlier in the topic \textit{Sustainable transport}, with .45 topic probability (Figure \ref{fig:violin_outliers}).

\paragraph{Co-occurrence in clusters across dimensions}
We have analyzed countries' co-occurrences in the same clusters. Three pairs of countries are in the same groups in all five dimensions mentioned earlier: Netherlands and Austria, Malta and Cyprus, Estonia and Finland. It may suggest that these countries frame issues similarly in all areas. France seems to be the most isolated country, co-occurrence only once with Belgium, Spain, Romania, and Slovenia. One reason explaining this could be that France has the most nuclear energy in their energy mix among other countries. 

We have also checked some insights regarding countries' alliances, e.g. the Visegrad Group, consisting of Czechia, Hungary, Poland, and Slovakia (Figure \ref{fig:visegrad}). In our analysis, there is one dimension, \textit{Energy efficiency}, in which all countries belong to the same cluster. In others, they are split into two clusters. What is worth seeing is that Poland and Hungary jointly appear in \textit{Decarbonization}, \textit{Energy efficiency}, and \textit{Energy security} dimensions. Czechia and Slovakia are together in \textit{Decarbonization}, \textit{Energy efficiency}, \textit{R\&I and Competitiveness}. 

Another interesting group is the Nordic countries (Figure \ref{fig:nordic}). In \textit{Internal market} all countries are in one cluster. In other categories, Denmark and Finland differ only in \textit{R\&I and Competitiveness}.
Figure \ref{fig:coocurencies} shows the full co-occurrence matrix.
\section{Conclusion}

Climate change is one of the most critical global threats. Creating public policies regarding this matter is essential in counteracting this problem. However, the creation of the policy itself is insufficient without the actions undertaken by the citizens and policymakers as a result. This is why our system allows for quick analysis of the focus of each national plan, comparing multiple countries across different dimensions, and detecting alignment between countries in policy framing. We believe that this could benefit the public understanding of the landscape of climate plans in the EU. 

Our method and pipeline can also be used to analyze other kinds of similarly structured policy documents, not only related to climate change. One of the documents that can be run through our framework in the future is a progress report produced by each Member State every two years regarding the changes they introduce as a result of their NECP. 

This work can be seen as a first step in developing methods for automated analysis of climate policies. In the future, such methods could allow tracking of how those policies change over time and how they relate to changes in the economic factors of countries, such as the energy mix.


\begin{ack}
We would like to thank Zuzanna Kwiatkowska for valuable discussions on the structure of the pipeline and the way of communicating the results and Ibrahim El-chami for mentoring and helpful comments.
Research was funded by (POB Cybersecurity and Data Science) of Warsaw University of Technology within the Excellence Initiative: Research University (IDUB) programme.
This research was carried out with the support of the Laboratory of Bioinformatics and Computational Genomics and the High Performance Computing Center of the Faculty of Mathematics and Information Science Warsaw University of Technology under computational grant number A-22-10.
\end{ack}

\bibliographystyle{plain}
\bibliography{references}

\clearpage
\appendix
\section{Number of public policies regarding climate change}
\begin{figure}[h]
\centering
\includegraphics[width=0.6\textwidth]{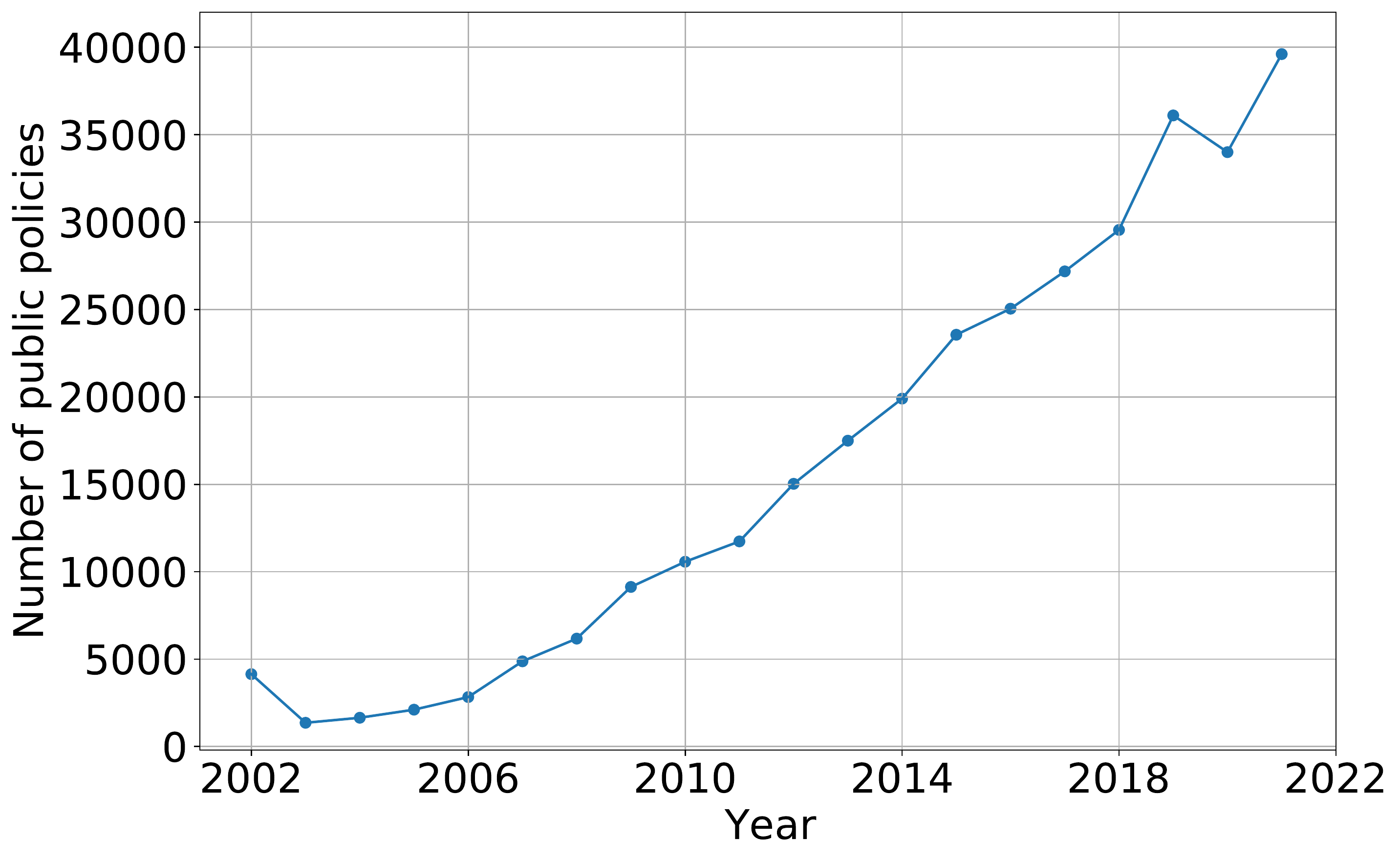}\\
\caption{Number of public policies regarding climate change in the \textit{Overton.io} database \cite{overton}.}
\label{fig:policies}
\end{figure}

There are multiple databases containing public policies. The biggest of them is \textit{Overton.io} database \cite{overton}. We have used query \textit{climate change} to check the number of public policies available. We got 322012 documents from 171 countries. Most of them was published by USA (81643), Intergovernmental Organizations (61884), United Kingdom (55393), and EU (40480). We have counted policies published each year, the results are shown in Figure \ref{fig:policies}. As we can see, the number of documents is rapidly growing.

\label{appendix:policies}

\section{Pipeline structure}
\label{appendix:pipeline}
\begin{figure}[h]
\centering
\includegraphics[width=\textwidth]{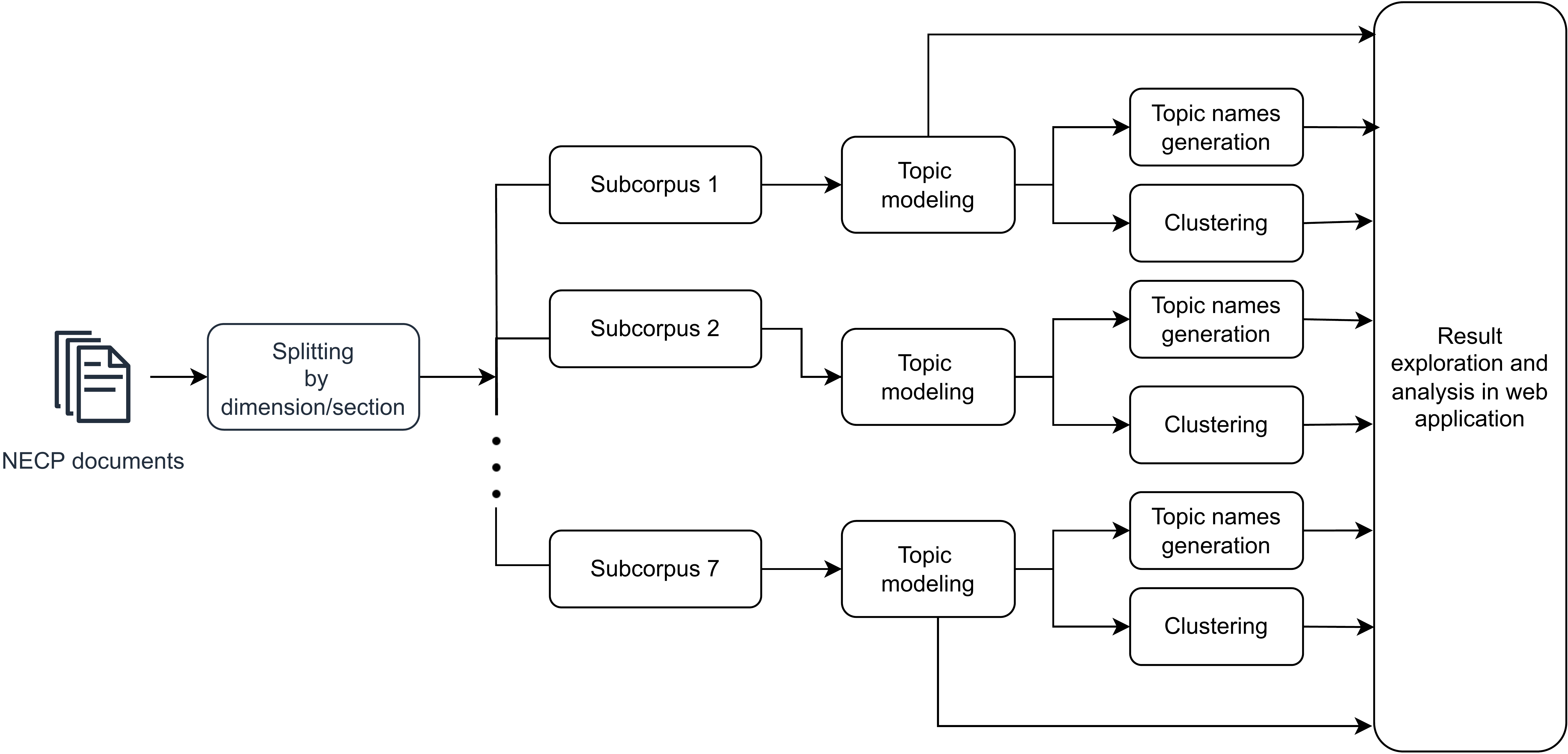}\\
\caption{Diagram of the proposed pipeline for automated analysis of national energy and climate plans with the use of  Natural Language Processing tools.}
\label{fig:pipeline}
\end{figure}

We define the proposed pipeline as a set of procedures and methods that are used to generate results, ranging from text extraction from documents to processing of data from trained models that can be explored using domain knowledge and analyzed using a web application. A simplified diagram of our \textit{Climate Policy Radar} pipeline is presented in Figure \ref{fig:pipeline}.

\section{Structure of National energy and climate plans}
\label{appendix:necp_structure}

\begin{figure}[h]
\centering
\includegraphics[width=\textwidth]{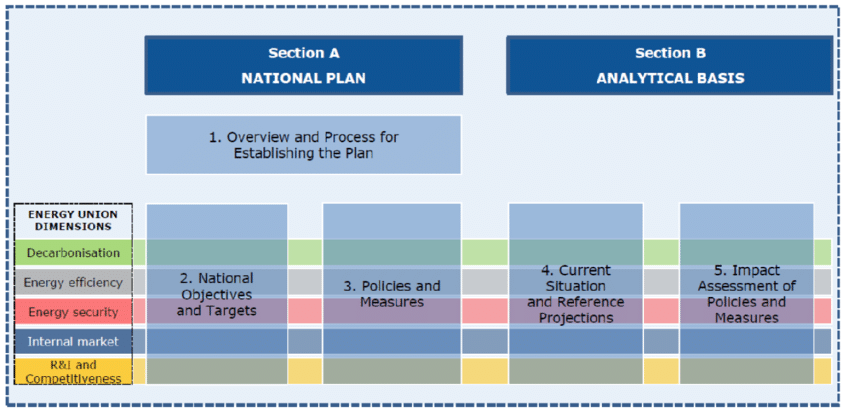}\\
\caption{Structure of~National Energy and~Climate Plans provided by the~European Commission.}
\label{fig:structure}
\end{figure}

The creation of a template for National energy and climate plans has been guaranteed by the Regulation of the European Parliament and of the Council on the Governance of the Energy Union and Climate Action \cite{regulation}. The diagram of the structure of the documents is presented in Figure \ref{fig:structure}. 

There are five Energy Union dimensions:
\begin{enumerate}
    \item \textit{Decarbonisation},
    \item \textit{Energy efficiency},
    \item \textit{Energy security}, 
    \item \textit{Internal market}, 
    \item \textit{R\&I and~Competitiveness}. 
\end{enumerate}

There are also five sections: 
\begin{enumerate}
\item \textit{Overview and~Process for Establishing the~Plan},
\item \textit{National Objectives and~Targets},
\item \textit{Policies and~Measures},
\item \textit{Current Situation and~Reference Projections},
\item \textit{Impact Assessment of~Planned Policies and~Measures}.
\end{enumerate}

The dimensions are well separated in sections (2)-(4), while the remaining sections constitute separate entities. It should be noted that there are discrepancies in the documents in relation to this structure. In particular, in the \textit{Impact Assessment of Policies and Measures} section, dimensions have not been explicitly separated.

\clearpage
\section{Text preparation}
\label{appendix:dataprep}
We used a well-defined structure of documents to create the corpus. First, we created a script that extracts the texts based on the table of contents. Then, since not every country strictly adhered to~the~imposed structure, we manually validated the results and corrected errors in problematic files. Ultimately, by creating separate records for each dimension in~each subsection, the~corpus of~453 documents were constructed. We applied a~set of~preprocessing steps to~the~prepared corpus with tools and~models available in~the~\texttt{spaCy} package \cite{spacy2}. We performed tokenization, lemmatization and added bigrams and~trigrams to unigrams with the aim to~improve the~topic modeling and~streamline topic interpretation.

\section{Web application for exploration of the results}
\label{appendix:app}
We present the process of examining the results obtained for \textit{Decarbonisation} dimension since it is, on average, the longest and most comprehensive part of the documents. For this dimension, we obtained five notably contrasting main topics presented in Table \ref{tab:topics} along with the corresponding keywords.

\begin{table}[hb!]
\centering
\caption{Topics and keywords for the \textit{Decarbonisation} dimension.}
\label{tab:topics}
\begin{tabular}{cc}
\toprule
\textbf{Topic name}      & \textbf{Keywords}                                                                                                                          \\ \midrule
\textit{Sustainable transport}    & \begin{tabular}[c]{@{}c@{}}support, plan, transport, vehicle, government, reduce, \\ public transport, car, green,   mobility\end{tabular} \\
\textit{National energy policy}   & \begin{tabular}[c]{@{}c@{}}promote, plan, national, system, integrated national energy,\\ policy measure, objective\end{tabular}           \\
\textit{All developments}         & \begin{tabular}[c]{@{}c@{}}development, electricity, preparation, production, implementation,\\  research development\end{tabular}         \\
\textit{GHG emissions}           & \begin{tabular}[c]{@{}c@{}}total, GHG emission, projection, gross final consumption,\\ share, GHG, decrease\end{tabular}                   \\
\textit{Renewable energy sources} & \begin{tabular}[c]{@{}c@{}}target, biomass, wind, solar, heat pump, bioenergy,\\ geothermal,  trajectory\end{tabular}     \\\bottomrule                
\end{tabular}                                               
\end{table}

\begin{figure}[h]
\centering
\begin{subfigure}{.6\textwidth}
  \centering
\includegraphics[width=\textwidth]{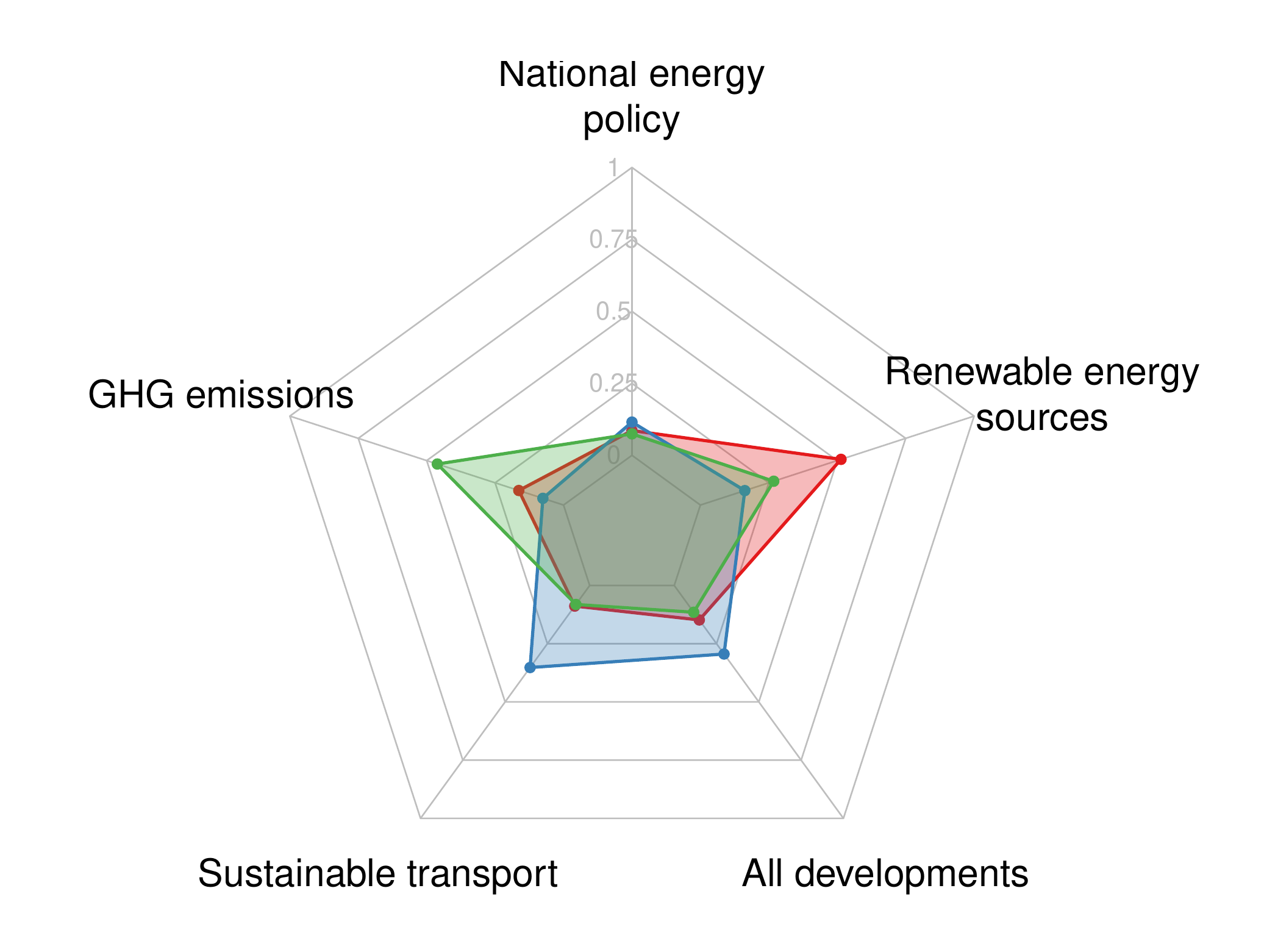}\\
\end{subfigure}%
\begin{subfigure}{.4\textwidth}
  \centering
\includegraphics[width=0.8\textwidth]{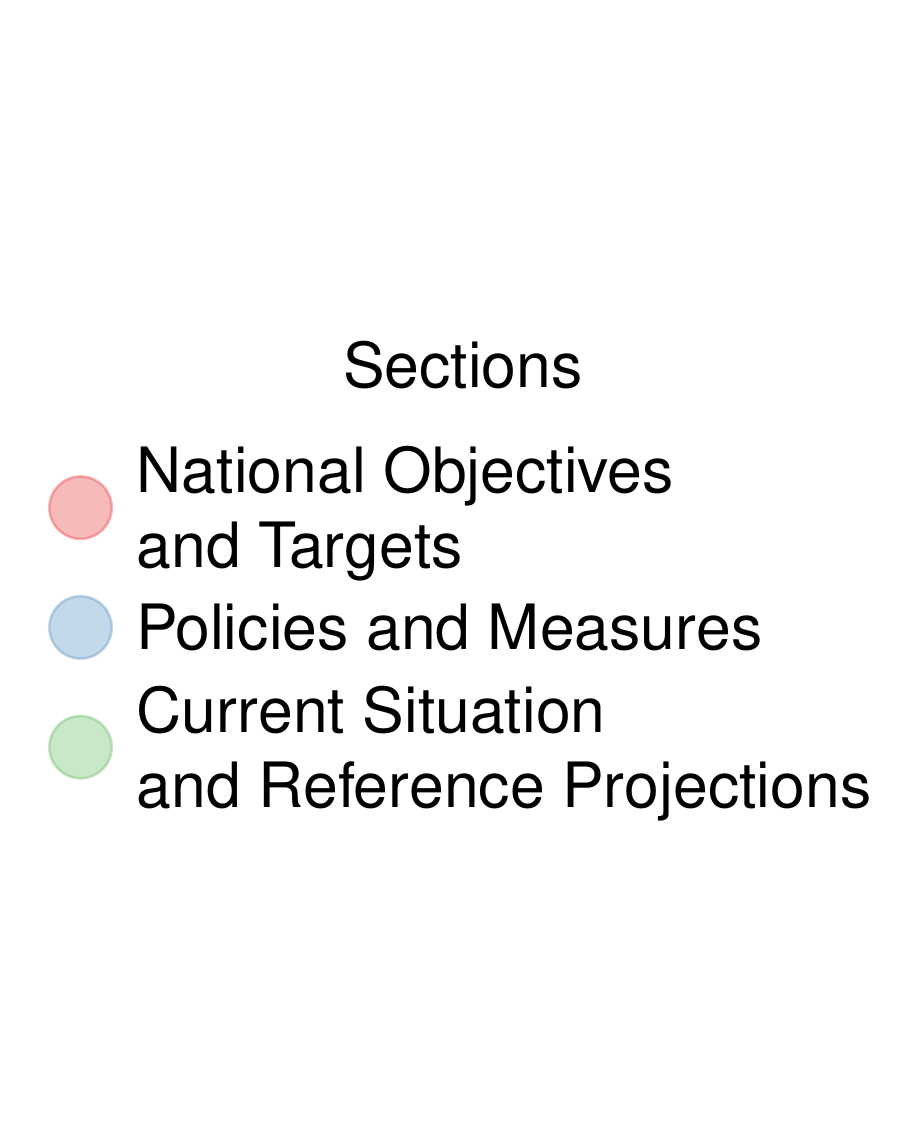}
\end{subfigure}
\caption{Topic distributions by section in the \textit{Decarbonisation} dimension for Finland.}
\label{fig:radar}
\end{figure}

\begin{figure}[h]
\centering
\includegraphics[width=\textwidth]{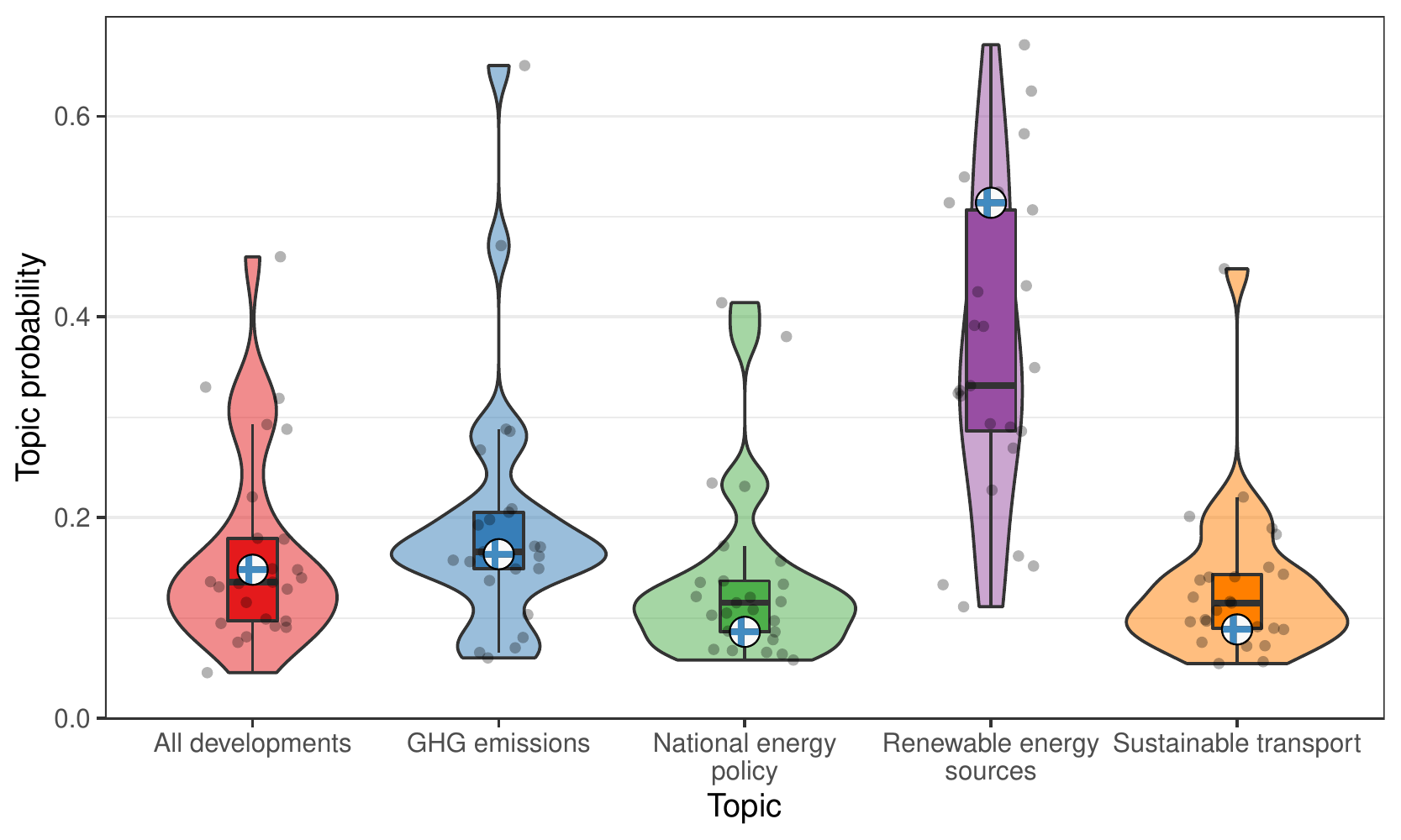}\\
\caption{Finland's topic distribution compared to other countries in the \textit{Decarbonisation} dimension in \textit{National Objectives and Targets} section.}
\label{fig:violin}
\end{figure}

The web application enables a detailed analysis of the results for a single selected Member State. Examples of visualizations facilitating the analysis are depicted in Figures \ref{fig:radar} and \ref{fig:violin} for Finland.

\newpage

\begin{figure}[h]
\centering
\includegraphics[width=\textwidth]{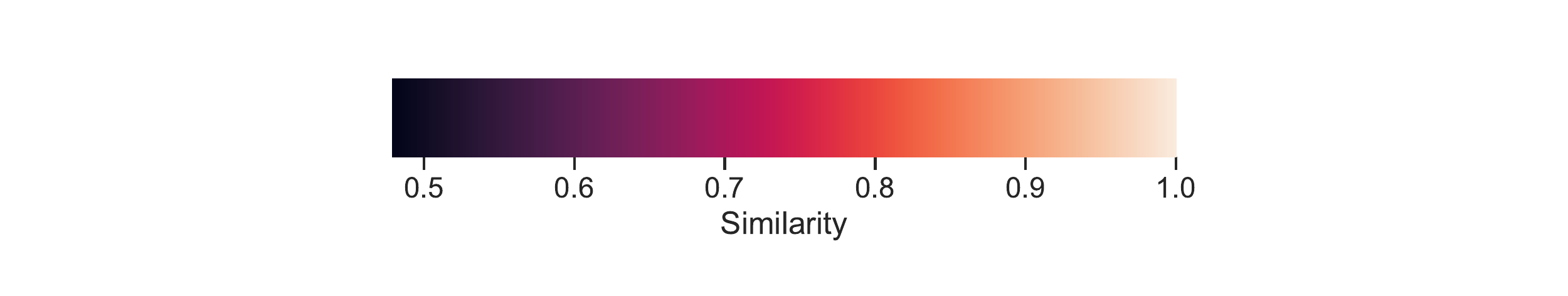}\\
\includegraphics[width=\textwidth]{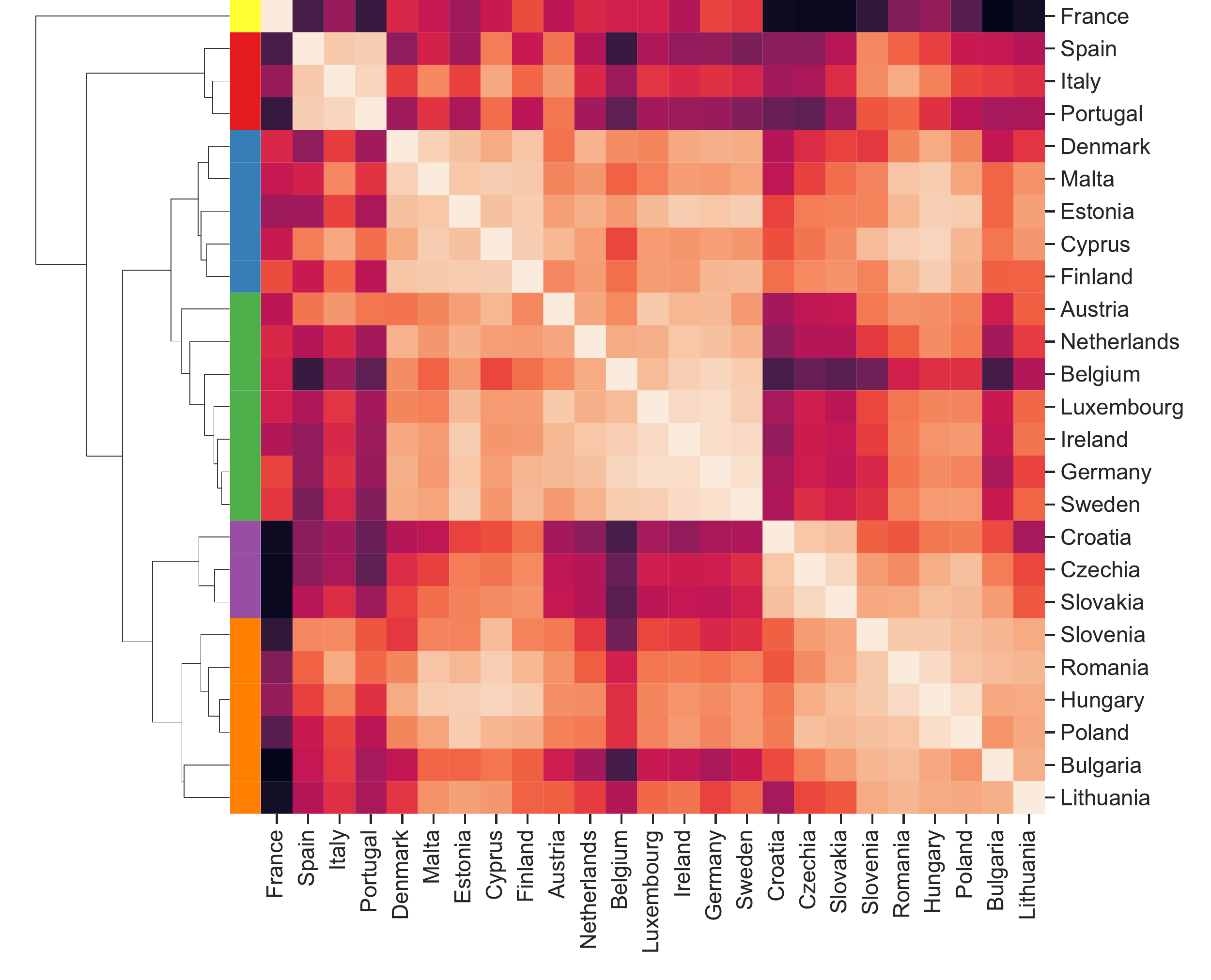}\\
\caption{Matrix of similarities between the bound distributions of topics for the \textit{Decarbonization} dimension. On the left side of the matrix, there is a dendrogram that visualizes the result of a hierarchical clustering calculation. The colors in the column preceding the matrix indicate the clusters in which individual countries appear on the basis of the selected distance threshold (country names are in the rows), see Figure \ref{fig:clustering}.}
\label{fig:hierarchical_clustering}
\end{figure}

The web application also allows for grouping countries with similar topic distributions in a selected dimension or section. Hierarchical clustering (selected for analysis in this study), HDBSCAN, or K-means algorithm can be used for this purpose. For dimensions (covered in three sections), clusterings are based on bound and normalized topic distributions. The result for the \textit{Decarbonization} dimension in the form country grouping is shown in Figure \ref{fig:clustering}. Distance threshold implying specific grouping was selected on the basis of the analysis of dendrogram and matrix of similarities between countries' agendas which are presented in Figure \ref{fig:hierarchical_clustering}. 


\clearpage

One of the key system features is the possibility of comparing the obtained topics (policy frames) with external variables in order to detect/clarify the dependencies. We examined the correlations with objectives, targets, and contributions under the Governance Regulation \cite{regulation}. 

The values were drawn from assessments of NECPs made by the European Commission. National targets and contributions values are precisely specified for 2020 and 2030. There are five different distinguished categories:
\setlist{nolistsep}
\begin{itemize}
    \item \texttt{greenhouse} -- a binding target for greenhouse gas emissions reductions compared to 2005 under the Effort Sharing Regulation (ESR);
    \item \texttt{renewable energy} -- national target/contribution for renewable energy: share of energy from renewable sources in the gross final consumption of energy;
    \item \texttt{primary energy} -- national contribution for energy efficiency: primary energy consumption;
    \item \texttt{final energy} -- national contribution for energy efficiency: final energy consumption;
    \item \texttt{electricity} -- level of electricity interconnectivity.

\end{itemize}

\begin{figure}[h]
\centering
\includegraphics[width=\textwidth]{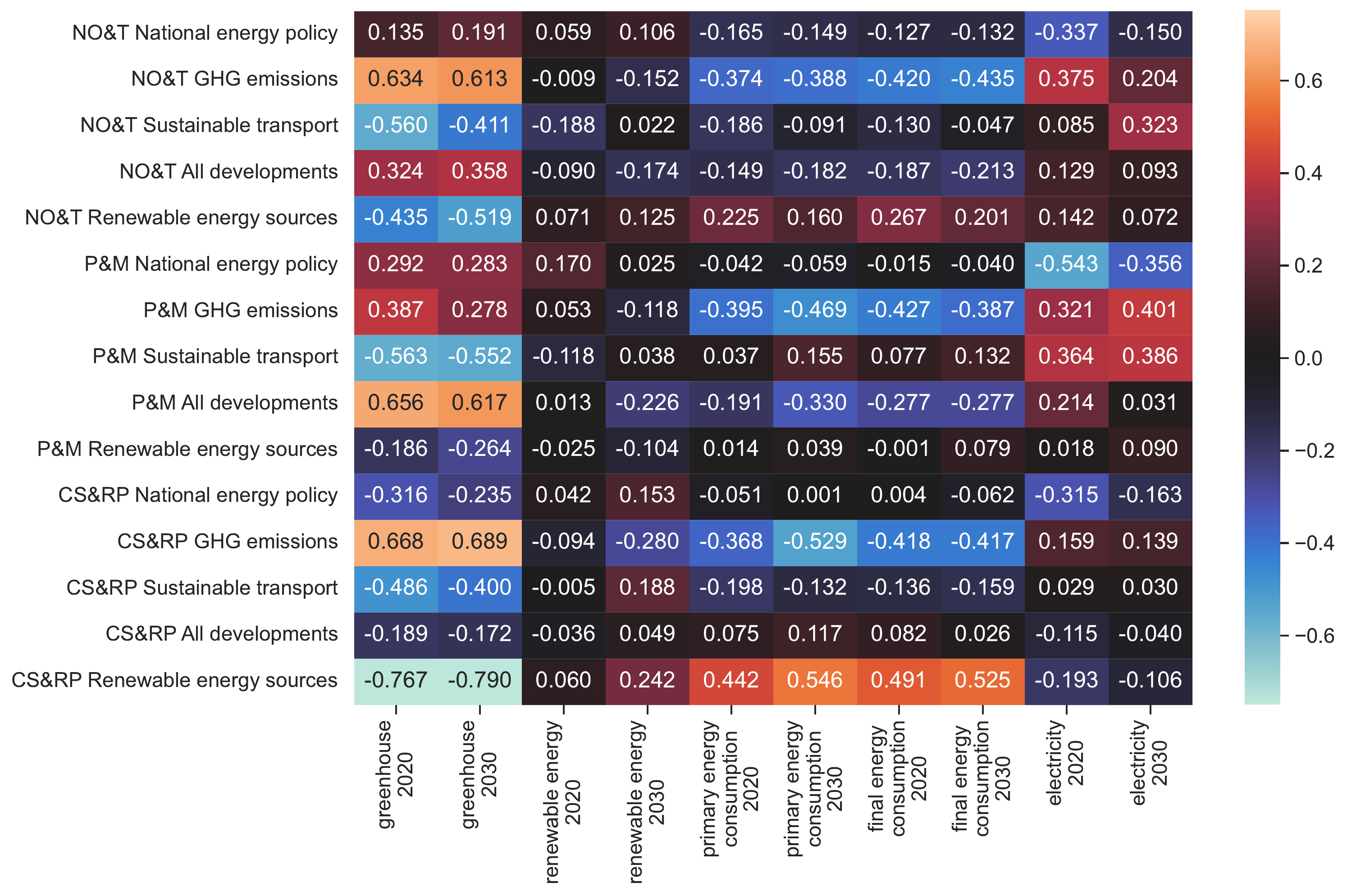}\\
\caption{Spearman correlations between topic distributions for the \textit{Decarbonisation} dimension in three sections (vertical axis; the section names are abbreviated for simplicity) and objectives extracted from the European Commission’s individual assessments of NECPs (horizontal axis).}
\label{fig:compare}
\end{figure}

\clearpage
\section{Policy frames}
\label{appendix:policyframes}

In the process of topic modeling, we have obtained the topics that may be seen as policy frames for Energy Union dimensions (Table \ref{tab:topics_dim}).

\setlist{nolistsep}
\begin{table}[h]
\begin{center}
\caption{Topics obtained for each Energy Union dimension.}
\label{tab:topics_dim}
\begin{tabularx}{\textwidth}[t]{XX}
\toprule
\textbf{Dimension} & \textbf{Topics}  \\
\midrule
\textit{Decarbonisation} & 
\begin{minipage}[t]{\linewidth}%
\begin{itemize}[leftmargin=*]
\item[1.1] Sustainable transport
\item[1.2] GHG emissions
\item[1.3] National energy policy
\item[1.4] Renewable energy sources
\item[1.5] All developments
\end{itemize} 
\end{minipage}\\

\midrule

\textit{Energy efficiency} & 
\begin{minipage}[t]{\linewidth}%
\begin{itemize}[leftmargin=*]
\item[2.1] Financial aspects
\item[2.2] Energy efficiency scenarios
\item[2.3] Saving energy
\item[2.4] Energy renovation
\item[2.5] Primary and final energy
\end{itemize} 
\end{minipage}\\

\midrule

\textit{Energy security} &
\begin{minipage}[t]{\linewidth}%
\begin{itemize}[leftmargin=*]
\item[3.1] Fossil fuels supply chain
\item[3.2] Energy import
\item[3.3] Capacity market development
\end{itemize}
\end{minipage}\\

\midrule

\textit{Internal market} &
\begin{minipage}[t]{\linewidth}%
\begin{itemize}[leftmargin=*]
\item[4.1] Natural gas
\item[4.2] Development and support
\item[4.3] Energy poverty alleviation
\end{itemize}
\end{minipage}\\

\midrule

\textit{R\&I and Competitiveness} &
\begin{minipage}[t]{\linewidth}%
\begin{itemize}[leftmargin=*]
\item[5.1] Research innovations
\item[5.2] Financial aspects
\item[5.3] Initiatives and programmes
\end{itemize}
\end{minipage}\\
\bottomrule

\end{tabularx}
\end{center}
\end{table}

Topic names for the final analysis were given manually, based on obtained keywords (see Appendix \ref{appendix:app}). Getting those topics was essential to see in which areas countries differ. The plot below shows an example of such analysis for the \textit{Decarbonisation} dimension.

\begin{figure}[h]
\centering
\includegraphics[width=0.9\textwidth]{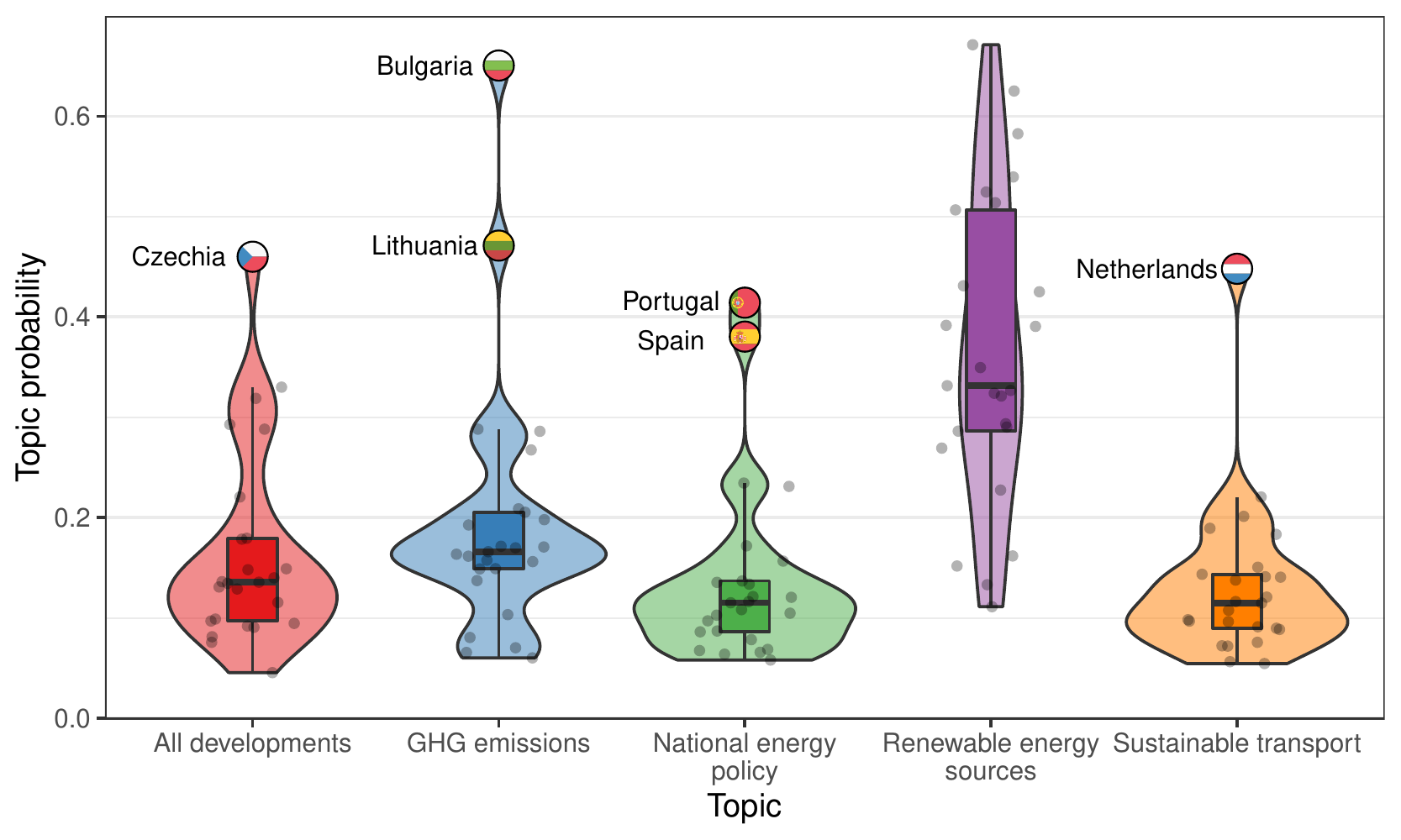}\\
\caption{Topic distributions for countries in the \textit{Decarbonisation} dimension in the \textit{National Objectives and Targets} section. Countries that could be recognized as outliers are marked with their flags and captioned.}
\label{fig:violin_outliers}
\end{figure}

\clearpage
\section{Differences between sections across dimensions}
\label{appendix:inconsistencies}

For the five Energy Union dimensions mentioned in all sections, we check the differences in the policy agendas for individual countries between the \textit{National Objectives and Targets} and \textit{Policies and Measures} sections. This analysis aims to show that there are notable disparities between the sections describing objectives and policies intended for achieving them. We calculate dissimilarities using the information radius measure applied to pairs of topic distributions for the corresponding documents from the corpus. For each dimension, we divide the obtained measures by the number of policy frames to account for their variable cardinality. 

We can see that there are countries where these differences are more pronounced, e.g. dissimilarities for Belgium in 4 out of 5 dimensions are higher than the average for all countries. The dimension with the greatest inconsistencies between the sections is \textit{Decarbonization}. On the other hand, on average, framing policies is the most consistent in the area related to \textit{R\&I and Competitiveness}.

\begin{figure}[h]
\centering
\includegraphics[width=0.9\textwidth]{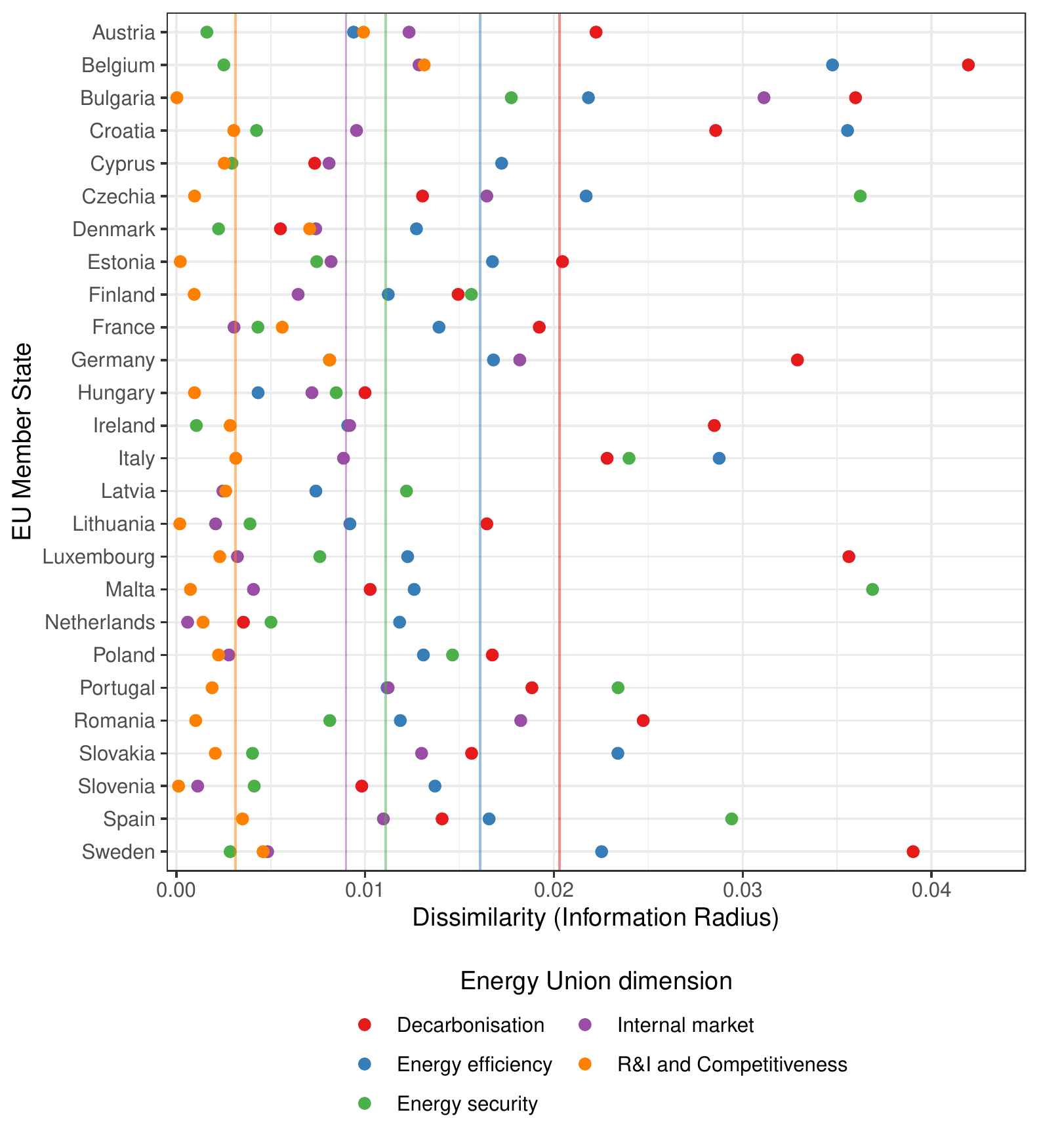}\\
\caption{Normalized dissimilarities (expressed as an information radius) between agendas of the \textit{Objectives and Targets} and \textit{Policies and Measures} sections for all countries and Energy Union dimensions. The greater the dissimilarity value, the less homogeneity there is between the topics covered in those sections. The colored lines show the average distance values for each dimension. }
\label{fig:inconsistencies}
\end{figure}

\clearpage
\section{Co-occurrence in clusters across dimensions}

\begin{figure}[h]
\centering
\includegraphics[width=\textwidth]{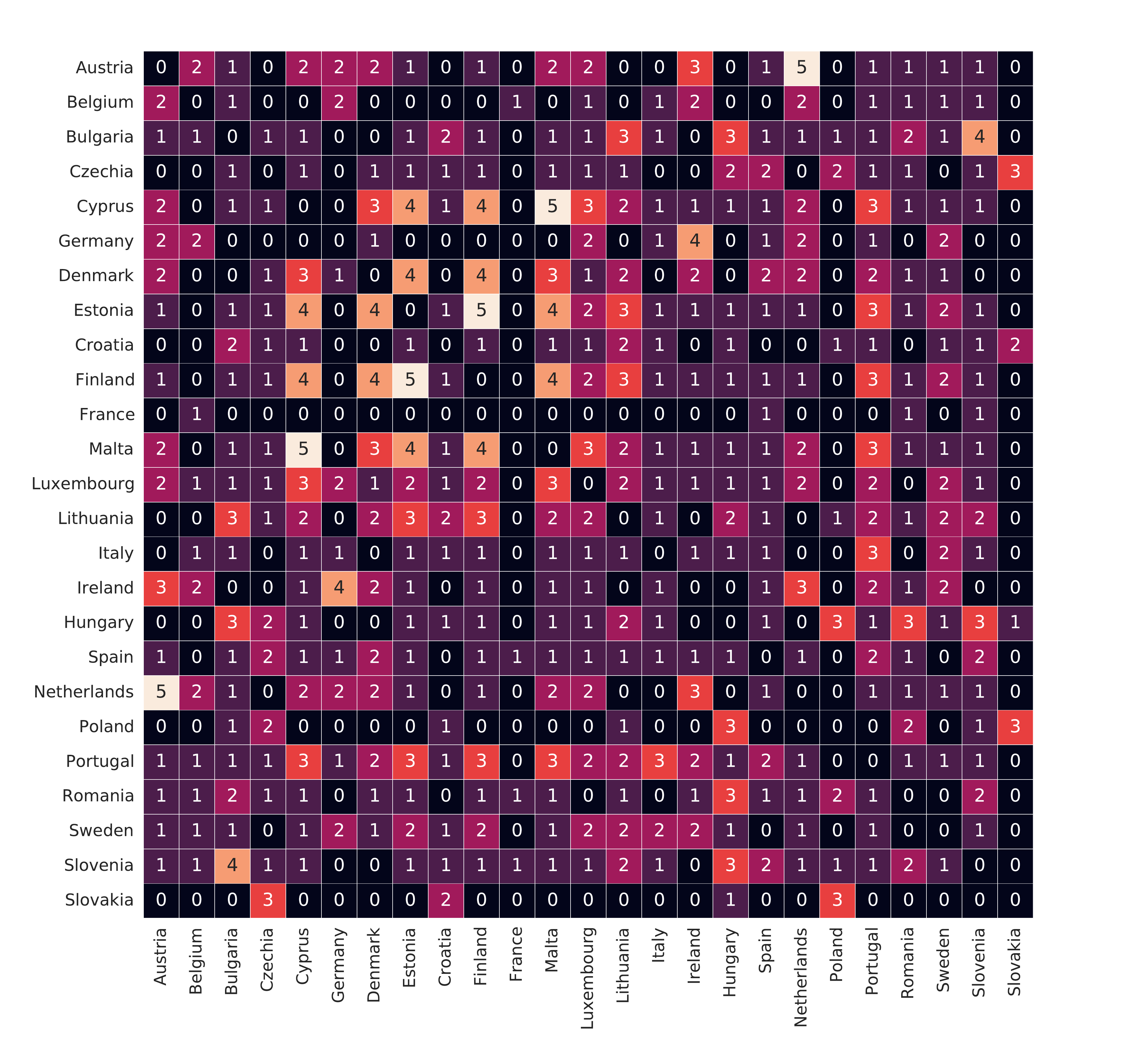}\\
\caption{Matrix with a number of co-occurrences of countries in the same clusters across clusterings for five Energy Union dimensions. We see several Member States being clustered together in all the dimensions, which suggests that they discuss different policy frames similarly.}
\label{fig:coocurencies}
\end{figure}

For all countries, we have checked which countries are together across all dimensions; by doing that, we wanted to check which countries often group. The result of this analysis is shown in Figure \ref{fig:coocurencies}. 

\begin{figure}[h]
\centering
\begin{subfigure}{.5\textwidth}
  \centering
  \includegraphics[width=\textwidth]{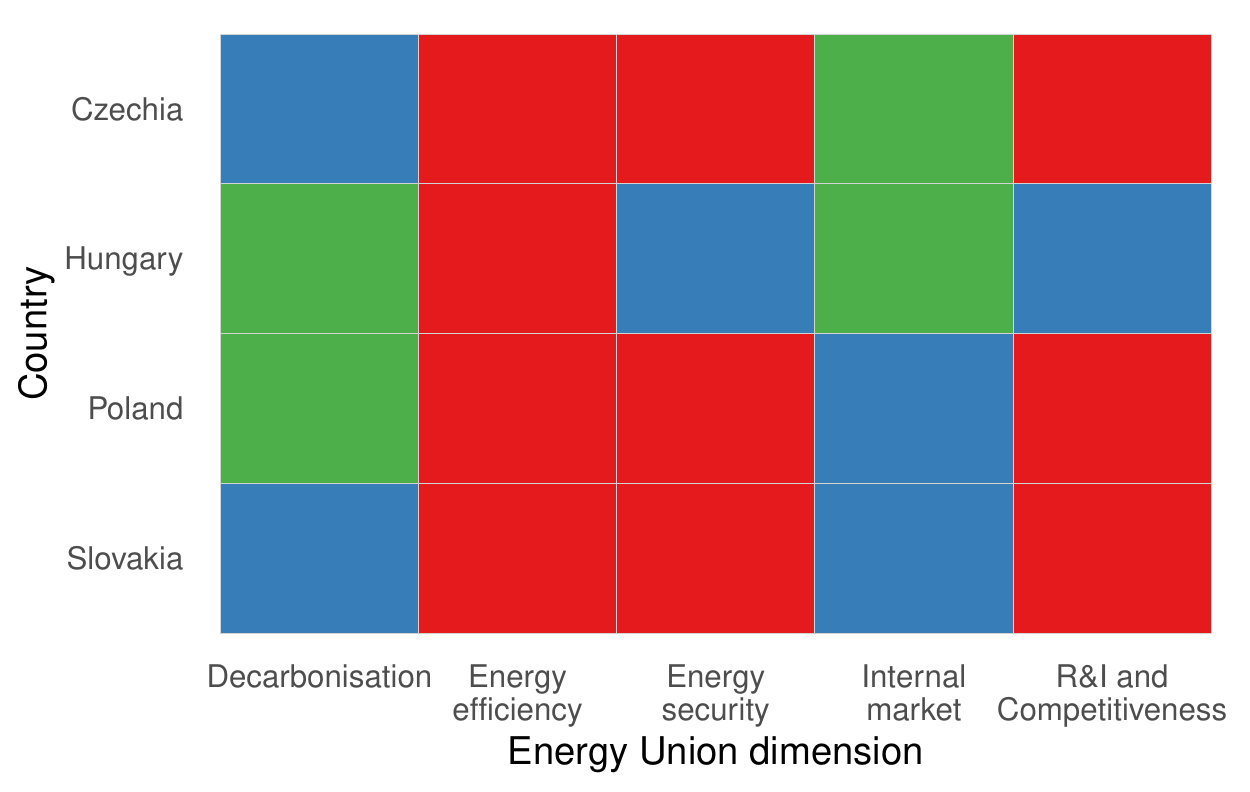}
  \caption{Assignment of countries from the Visegrad Group.}
  \label{fig:visegrad}
\end{subfigure}%
\begin{subfigure}{.5\textwidth}
  \centering
  \includegraphics[width=\textwidth]{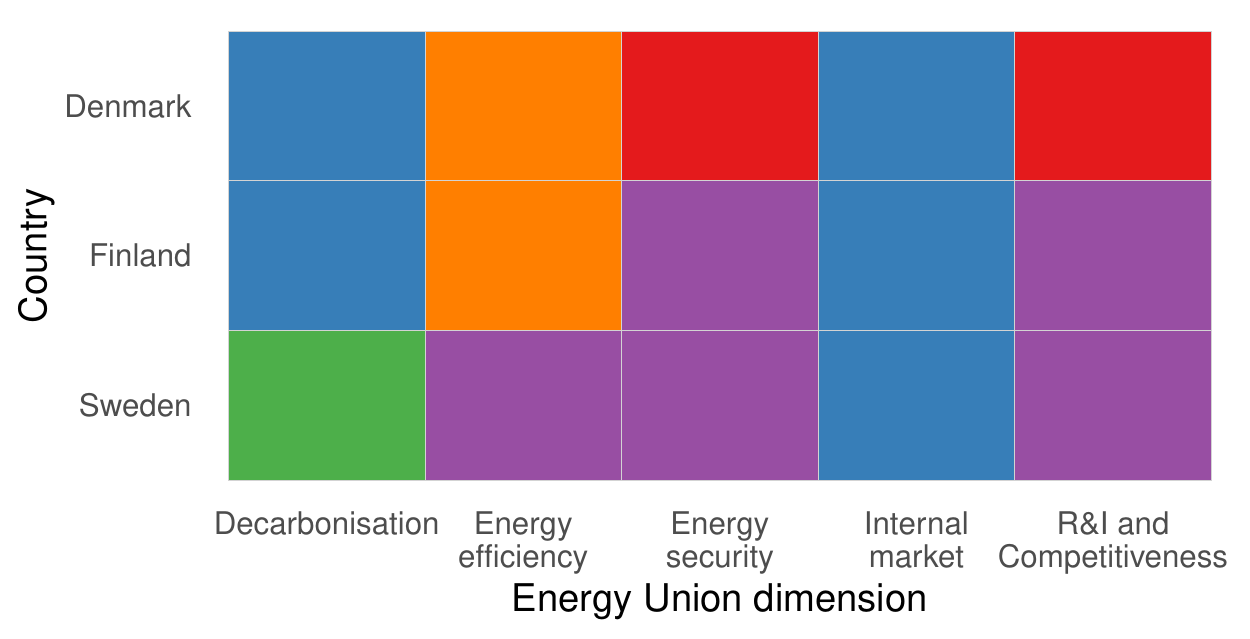}
  \caption{Assignment of EU Nordic countries. }
  \label{fig:nordic}
\end{subfigure}
\caption{Assignment of countries to clusters in clustering obtained for five Energy Union dimensions. Each color means being together in one cluster.}
\label{fig:groups}
\end{figure}

We have also checked some known groups: the Visegrad Group and Nordic countries, to check if some political alliances or geographical location factor into clustering. As we can see in Figure \ref{fig:groups}, there are visible similarities in clusters in both cases.

\end{document}